\def\year{2022}\relax
\documentclass[letterpaper]{article} % DO NOT CHANGE THIS
\usepackage{aaai22}  % DO NOT CHANGE THIS
\usepackage{times}  % DO NOT CHANGE THIS
\usepackage{helvet}  % DO NOT CHANGE THIS
\usepackage{courier}  % DO NOT CHANGE THIS
\usepackage[hyphens]{url}  % DO NOT CHANGE THIS
\usepackage{graphicx} % DO NOT CHANGE THIS
\usepackage{natbib}  % DO NOT CHANGE THIS AND DO NOT ADD ANY OPTIONS TO IT
\usepackage{caption} % DO NOT CHANGE THIS AND DO NOT ADD ANY OPTIONS TO IT
\DeclareCaptionStyle{ruled}{labelfont=normalfont,labelsep=colon,strut=off} % DO NOT CHANGE THIS
\usepackage{algorithm}
\usepackage{algorithmic}

\usepackage{graphicx}
\usepackage{enumitem}
\usepackage{amsmath}
\usepackage{subfigure}
\usepackage{booktabs}
\usepackage{multirow}
\usepackage{amssymb}
\usepackage{hyperref}
\usepackage{stmaryrd}

\usepackage{newfloat}
\usepackage{listings}
\floatstyle{ruled}
\newfloat{listing}{tb}{lst}{}
\floatname{listing}{Listing}
\title{AAAI Press Formatting Instructions \\for Authors Using \LaTeX{} --- A Guide}
\title{Learning to Ask for Data-Efficient Event Argument Extraction}

\author {
        Hongbin Ye$^{1,2}\footnote{Equal contribution and shared co-first authorship.}$,
        Ningyu Zhang$^{1,2*\dagger}$,
        Zhen Bi$^{1,2}$,
        Shumin Deng$^{1,2}$,
        Chuanqi Tan$^{3}$,\\
         Hui Chen$^{3}$,
        Fei Huang$^{3}$,
        Huajun Chen$^{1,2}\footnote{Corresponding author.}$
        \\
}
\affiliations {
    \textsuperscript{\rm 1} Zhejiang University \& AZFT Joint Lab for Knowledge Engine, No.38 Zheda Road, Hangzhou, +86 13605851726\\
    \textsuperscript{\rm 2} Hangzhou Innovation Center, Zhejiang University,
    \textsuperscript{\rm 3} Alibaba Group \\ 
 \texttt{ \{yehongbin,zhangningyu,bizhen\_zju,231sm,huajunsir\}@zju.edu.cn} \\ 
      \texttt{\{chuanqi.tcq,weidu.ch,f.huang\}@alibaba-inc.com}
    
}

\usepackage{bibentry}
\begin{document}

\maketitle

\begin{abstract}
Event argument extraction (EAE) is an important task for information extraction to discover specific argument roles. In this study, we cast EAE as a question-based cloze task and empirically analyze fixed discrete token template performance. As generating human-annotated question templates is often time-consuming and labor-intensive, we further propose a novel approach called “Learning to Ask,” which can learn optimized question templates for EAE without human annotations. Experiments using the ACE-2005 dataset demonstrate that our method based on optimized questions achieves state-of-the-art performance in both the few-shot and supervised settings.
\end{abstract}

\section{Introduction}

Event argument extraction (EAE) is an important and challenging task in information extraction, which aims to discover specific role types for each argument in an event. 
For example, given that the word \textit{“declared bankruptcy”} triggers a Declare-Bankruptcy event in the sentence \textit{“My uncle declared bankruptcy in 2003 and his case closed in June 2004”}, EAE aims to identify that \emph{\textbf{“My uncle”}} is an event argument in this sentence, and its argument role is \emph{\textbf{“Org”}}. 
Previous methods of EAE have relied heavily on using a large volume of training data, which prohibits the use of these methods in low-data scenarios \cite{DBLP:conf/emnlp/LiuCLBL20}.

\begin{figure}[tb]
\centering
  \includegraphics[width=0.46\textwidth]{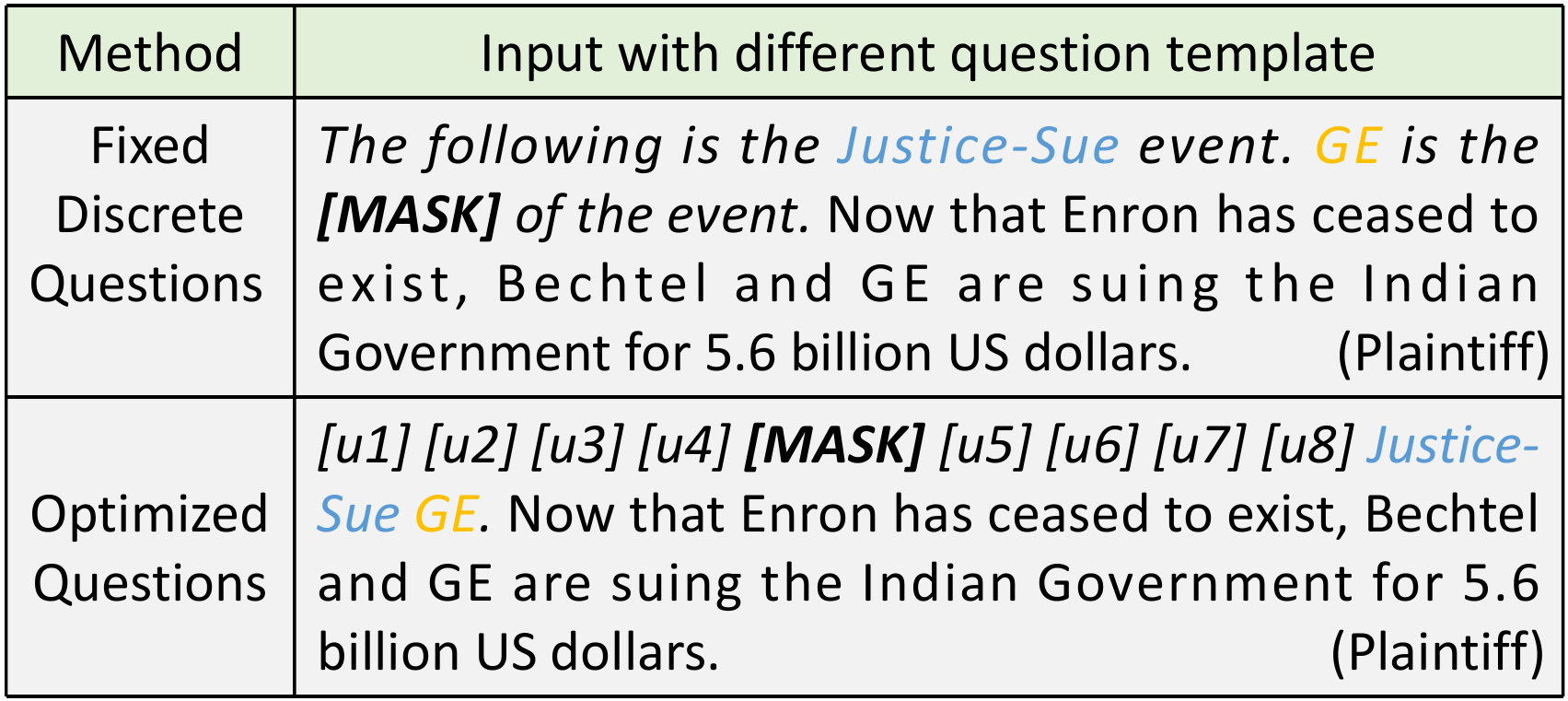}
\caption{Event argument extraction as a question-based cloze task, conducted in a masked language modeling manner. In the second example, we use the pseudo question token to search for the most probable argument  role type.}
\label{intro}
\end{figure}

Recently, Event Extraction (EE) has been reintroduced as a machine reading comprehension problem (MRC) 
% \cite{liu_event_2020,li_event_2020,du_event_2021}.
\cite{du_event_2021}.
Within this paradigm, question templates are used to map an input sentence to a suitable sequence to improve performance.  
For EAE, there is still a need to optimize an additional parameter matrix for classification, which is particularly challenging when the amount of data available is limited. 
To bridge the gap between fine-tuning and pre-training, inspired by the recently emerging prompt-based methodology \cite{chendanqi}, we take the first step to cast EAE as a question-based cloze task.
As shown in Figure \ref{intro}, we concatenate a question with the input sentence and leverage [MASK] to indicate the event type for subsequent predictions. 
It should be noted that we use declarative sentences as question templates because they can increase semantic consistency across contexts and improve prediction performance. As different strategies of question-asking can differently affect the performance of event extraction methods, we propose a novel method called Learning to Ask (\textbf{L2A}), to optimize questions with backpropagation, which can automatically search for the best pseudo question token in the continuous embedding space. 

\section{Our Model: L2A}
\label{modelpart}

Our L2A framework for EAE relies on question templates $X_{Q}$ that map an input sentence to a QA-formatted input sequence for a standard pre-trained bidirectional transformer: 
$
[\mathrm{CLS}] \textbf{question}[\mathrm{SEP}] \textbf{sentence}[\mathrm{SEP}]
$.
For the construction of the template for the \textbf{question},
we introduce two different strategies:
(1) \textbf{L2A (base)}: the manual questions template for the input text, which replace the tokens of the argument roles with the \textbf{[mask]} and add necessary prompt word information, such as event type and argument span tokens, into the question template. 
(2) \textbf{L2A (pseudo)}:
because manual prompts are labor-intensive and may result in sub-optimal EAE performance, we further introduce the automatic construction of the question template. 
Specifically, we use several unused tokens \textbf{[u1] - [u8]} (e.g., unused or special token in the vocabulary) to form a pseudo question template and fix the other weights of the language model to learn the optimized questions. 

\begin{table}[!t]\centering
\begin{tabular}{|l|ccc|}
\hline
\multirow{2}{*}{Models} & \multicolumn{3}{c|}{\begin{tabular}[c]{@{}c@{}}Argument Role\\ Classificaiton (\%)\end{tabular}} \\ \cline{2-4} 
                        & P                              & R                              & F                             \\ \hline
% Cross-Event            & 45.1                           & 44.1                           & 44.6                          \\
% Cross-Entity            & 51.6                           & 45.5                           & 48.3                          \\
% DMCNN                  & 62.2                           & 46.9                           & 53.5                          \\
% % RBPB                 & 41.9                           & 36.5                           & 39.0                          \\
% JRNN                & 54.2                           & 56.7                           & 55.4                          \\
% dbRNN                   & 66.2                           & 52.8                           & 58.7                          \\
JMEE                  & 66.8                           & 54.9                           & 60.3                          \\
% DMCNN-DS                & 62.8                           & 50.1                           & 55.7                          \\
% PLMEE                   & 62.3                           & 54.2                           & 58.0                          \\ 
HMEAE                  & 62.2                           & 56.6                           &  59.3  \\

RCEE                   & 63.0                           & 64.2                           &  63.6     \\ 
 
% BERT-QA                   & 56.77                           & 50.24                           & 53.31
BERT-QA                   & 56.7                           & 50.2                           & 53.3
\\\hline

L2A (base)                  &  66.9                          &  63.4                          & 65.1                         \\
L2A (pseudo)         & \textbf{69.3*}                               &  \textbf{66.9*}                              & \textbf{68.1*}                              \\

L2A (projected)                 & 68.5                               & 66.1                               & 67.3                              \\ \hline
\end{tabular}
\caption{Full supervision results using the ACE-2005 dataset.  
* denotes a significance level of $p = 0.05$.}
\label{table1}
\end{table}

\subsection{Training Objectives}

Because the argument role label contains semantic information, we can simplify the label mapping in EAE as an injective function. 
For example, we can project the \emph{"Transaction.Transfer-Money"} event role as: 
\begin{equation}
\mathcal{Y}(\mathcal{R}) = \{giver,recipient,beneficiary,place\}
\end{equation}
% , where $\mathcal{Y}(\mathcal{R})$ represents the role projection function of the event element of the $i^{th}$ event type. 
We normalize the vocabulary distribution of a single token of the event role and define the prediction probability as:
\begin{equation}
p\left(y \mid X_{\text {Q }}\right)=\frac{\exp \left(\mathbf{w}_{\mathcal{Y}(r)} \cdot \mathbf{h}_{[\text {MASK}]}\right)}{\sum_{r^{\prime} \in \mathcal{R}} \exp \left(\mathbf{w}_{\mathcal{Y}\left(r^{\prime}\right)} \cdot \mathbf{h}_{[\text {MASK}]}\right)}
\end{equation}
, where $\mathbf{h}_{[\text {MASK}]}$ is the hidden vector corresponding to the $[\text {MASK}]$ position.
Therefore, we use cross-entropy loss to define the event role prediction as:
\begin{equation}
\mathcal{L}_{EAE}=\operatorname{CE}\left(p\left(y \mid X_{\text {Q }}\right)\right)
\end{equation}
where $\mathcal{L}_{EAE}$ denotes the EAE loss and $CE$ is the cross entropy loss function. To make the input text resemble the natural language more closely, we leverage an auxiliary optimization object. We randomly mask out other tokens in the sentence and conduct a masked language model prediction as:
\begin{equation}
q\left(x^{m} \mid x^{\prime}, u\right)=\frac{\exp \left(\llbracket L \left(x^{\prime}, u\right) \rrbracket_{x^{m}}\right)}{\sum_{v^{\prime} \in \mathcal{V}} \exp \left(\llbracket L\left(x^{\prime}, u\right) \rrbracket_{v^{\prime}}\right)}
\end{equation}
\begin{equation}
\mathcal{L}_{MLM}=\sum_{m \in M} \operatorname{BCE}\left(q\left(x^{m} \mid x^{\prime}, u\right)\right)
\end{equation}
, where $u$ denotes the question format input sequence, $x^{m}$ is the original token $x$ which has been randomly masked, $x^{\prime}$ represents the input sentence after mask processing,
and $BCE$ is the binary cross entropy loss function. 
Finally, we have the following optimization object:
\begin{equation}
\mathcal{L}_{total}=\mathcal{L}_{EAE}+\mathcal{L}_{MLM} 
\end{equation}

% \section{model}
% L2A replaces the input embeddings of pre-trained language models with its own differentiable question embedding that is concatenated with the input embedding, which only applies the noninvasive modification to the input.
% Let $X_{in}=\{x_1, x_2, \ldots,X_t,\ldots, X_a, x_N\}$ be a sequence, where $x_n$ is the $n^{th}$ token in the sequence, and $N$ is the number of tokens. 
% The goal of EAE is to predict the argument role  $r \in \mathcal{R}$  given the trigger $X_t$ and candidate argument $X_a$ respectively. 
% %We convert $X_{in}$ to an embedding token sequence $\tilde{X}_{\mathrm{in}}$ and then map $\tilde{X}_{\mathrm{in}}$ to a sequence of hidden vectors $\left\{\mathbf{h}_{k} \in \mathbb{R}^{d}\right\}$. 
% We concatenate the question $u$ with the input sequence  $X_{in}$ as $X_{\text{Q}}$ and cast our task as a language modeling task: 
% % $
% % p\left([\mathrm{MASK}] \mid\left(X_{\text{Q }}\right)\right.$ 
% % Probability of an output argument role label is: 
% \begin{equation}
% p\left(y \mid X_{\text{Q}}\right)= p\left([\mathrm{MASK}]=r\mid X_{\text{Q}}\right)
% \end{equation}
% , where $r$ denotes the label of the argument type. This reduces the gap between fine-tuning and pretraining, making it more effective to use both in fully supervised and few-shot scenarios.

\section{Experiments}
\label{Experiments}

\begin{figure}[!t]
\centering
  \includegraphics[width=0.38\textwidth]{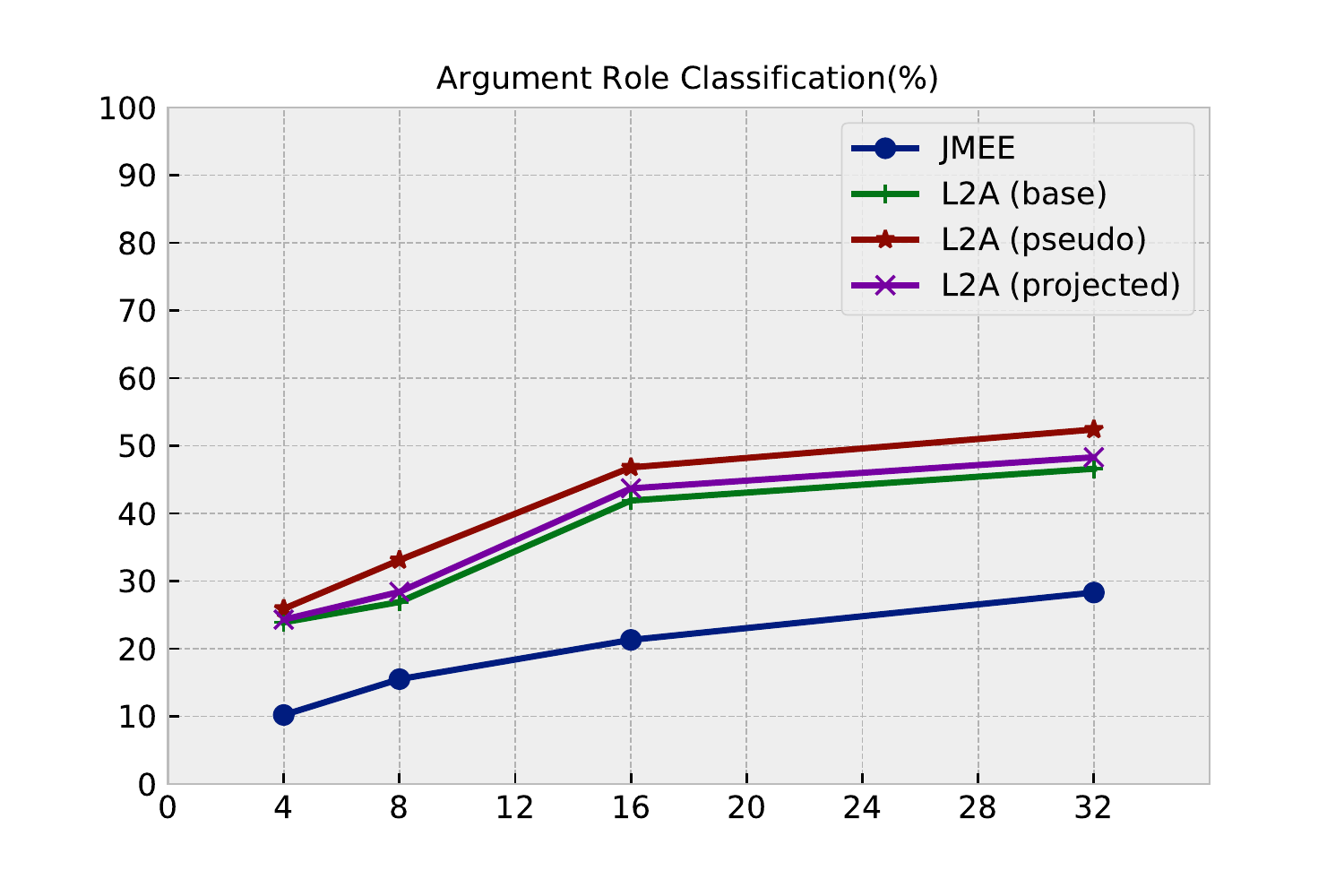}
\caption{Few-shot results (random sampling) using the ACE-2005 datasets. We use K = 4, 8, 16, 32 (\# examples per class) for few-shot experiments. The labels “base” and “pseudo” refer to the model with manual and optimized questions, respectively.}
 
\label{few}
\end{figure}

We evaluate our L2A model with the ACE 2005 dataset. For the few-shot scenario, we follow the few-shot settings of \cite{chendanqi}, which is different from the N-way K-shot setting. 
From Table \ref{table1}, we observe that L2A (base) performs better than baselines, which indicates that fine-tuning in a question-based cloze task can bring substantial benefits. 
To illustrate the effectiveness of optimizing tokens, we conduct a nearest neighbor vocabulary embedding search to project the best optimized pseudo question token to a readable natural language.
We note the model with the projected question as L2A (projected).
% \footnote{We selected the top 10 nearest neighbor tokens and combined them into sentences in the supplementary materials.}.
It is worth noting that the performance of L2A (projected) is only lower than the best-optimized result by 0.8\%.
From Figure \ref{few}, we can observe that L2A (pseudo) demonstrates an absolute improvement of up to \textbf{15\%} over previous state-of-the-art models in the few-shot scenario (k = 4).
Note that the question can include task-specific and argument-relevant information that can boost the performance. 
Moreover, our approach is consistent with the pre-training paradigm and thus is more convenient for leveraging the knowledge available in the parameter space when learning with sparse data. 

\section{Conclusion}

In this paper, we present L2A, a simple but effective method for optimizing question templates for EAE tasks, and demonstrate that it performs competitively with state-of-the-art models.
Without any human effort or annotations, L2A can search for suitable questions for EAE and obtain better performance in few-shot and fully supervised settings than previous methods, making it broadly applicable to a variety of real-world scenarios.

\bibliography{aaai22.bib}

\end{document}